\documentclass[10pt,twocolumn,letterpaper]{article}

\usepackage[pagenumbers]{cvpr}

\usepackage{graphicx}
\usepackage{amsmath}
\usepackage{amssymb}
\usepackage{booktabs}
\usepackage{multirow}
\usepackage{arydshln}
\usepackage{array}

\usepackage[pagebackref,breaklinks,colorlinks]{hyperref}
\usepackage[capitalize]{cleveref}
\crefname{section}{Sec.}{Secs.}
\Crefname{section}{Section}{Sections}
\Crefname{table}{Table}{Tables}
\crefname{table}{Tab.}{Tabs.}

\newcommand{\tablestyle}[2]{\setlength{\tabcolsep}{#1}\renewcommand{\arraystretch}{#2}\centering\footnotesize}

\def\cvprPaperID{1980} 
\def\confName{CVPR}
\def\confYear{2023}

\begin{document}

\title{A Zero-/Few-Shot Anomaly Classification and Segmentation Method for \\CVPR 2023 VAND Workshop Challenge Tracks 1\&2: \\1st Place on Zero-shot AD and 4th Place on Few-shot AD}
\author{Xuhai Chen$^1$\thanks{Equal contribution.}
~ ~ Yue Han$^{1}$\footnotemark[1]
~ ~ Jiangning Zhang$^{2}$\footnotemark[1]~~\thanks{Project Leader.}\\
\normalsize $^1$APRIL Lab, Zhejiang University ~ ~ $^2$Youtu Lab, Tencent \\
{\tt\small \{22232044, 22132041\}@zju.edu.cn, vtzhang@tencent.com} \\
}
\maketitle

\begin{abstract}
In this technical report, we briefly introduce our solution for the Zero/Few-shot Track of the Visual Anomaly and Novelty Detection (VAND) 2023 Challenge. For industrial visual inspection, building a single model that can be rapidly adapted to numerous categories without or with only a few normal reference images is a promising research direction. This is primarily because of the vast variety of the product types. For the zero-shot track, we propose a solution based on the CLIP model by adding extra linear layers. These layers are used to map the image features to the joint embedding space, so that they can compare with the text features to generate the anomaly maps. Besides, when the reference images are available, we utilize multiple memory banks to store their features and compare them with the features of the test images during the testing phase. In this challenge, our method achieved first place in the zero-shot track, especially excelling in segmentation with an impressive F1 score improvement of 0.0489 over the second-ranked participant. Furthermore, in the few-shot track, we secured the fourth position overall, with our classification F1 score  of 0.8687 ranking first among all participating teams\footnote{\url{https://github.com/ByChelsea/VAND-APRIL-GAN}}.
\end{abstract}

\section{Introduction}
\label{sec:introduction}
In the field of computer vision, unsupervised anomaly detection (AD)~\cite{pathcore, fastflow, padim, ocrgan, m3dm} aims to identify abnormal images and locate anomalous regions using a model trained solely on anomaly-free images. It is widely used in industrial defect detection~\cite{mvtecad, mvtecloco, btad}. Most previous methods have centered on training dedicated models for each category, relying on a vast collection of normal images as references~\cite{draem, mkd}. In real-world applications, however, there are a wide variety of industrial products to be detected, and it is difficult to collect a large number of training images for each category. As a result, the zero/few-shot setting plays a crucial role in bringing AD to practical applications.

WinCLIP~\cite{winclip}, built on the open-source vision-language model CLIP~\cite{clip, openclip}, is a model that can be rapidly adapted to abundant categories without or with only a handful of normal images. It assumes that language is able to aid zero/few-shot AD and proposes a window-based strategy to perform segmentation. Taking inspiration from it, we also followed the pattern of language guided AD and employed CLIP as our baseline.

Specifically, we adhere to the overall framework of CLIP for zero-shot classification and employ a combination of state and template ensembles to craft our text prompts. In order to locate the abnormal regions, we introduce extra linear layers to map the image features extracted from the CLIP image encoder to the linear space where the text features are located. Then, we make similarity comparison between the mapped image features and the text features, so as to obtain the corresponding anomaly maps. For the few-shot case, we retain the extra linear layers of the zero-shot phase and maintain their weights. In addition, we use the image encoder to extract the features of the reference images and save them to the memory bank for comparison with the features of the test images during the test phase. Note that, to fully take advantage of the shallow and deep features, we leverage features from different stages for both the zero and few shot settings.

Our method demonstrates strong competitiveness in both zero-shot and few-shot anomaly classification and segmentation. In the zero-shot track of the VAND Challenge, we achieve the first rank, surpassing the second-ranked approach by 0.0633 in overall score and 0.0489 in segmentation F1 score. In the few-shot track, we rank fourth overall, but our classification F1 score of 0.8687 places us first, surpassing the second-ranked approach by 0.0207.

\section{Methodology}
Our approach is based on the CLIP~\cite{clip} model, an open-source vision-language model that inherently possesses the capability for zero-shot classification. This implies that CLIP may be directly applicable to zero-shot anomaly classification. Furthermore, we made slight modifications to this model to enable its usage in anomaly segmentation and extended its applicability to the few-shot domain.

\begin{figure*}
  \centering
  \includegraphics[width=0.95\linewidth]{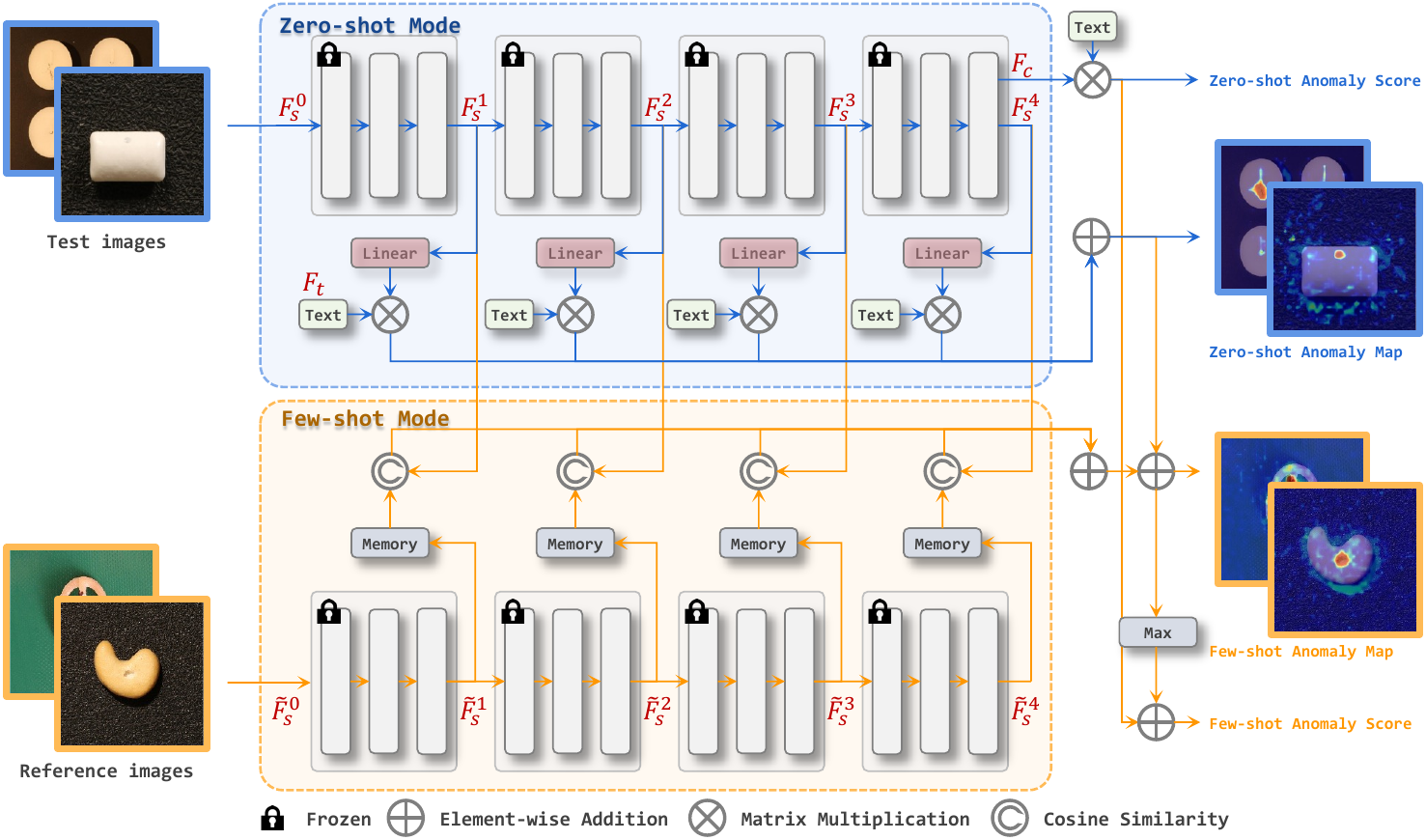}
  \caption{\textbf{Overall diagram of our solution}. \textbf{\textit{1)}} The blue dashed box represents the pipeline for the zero-shot settiing. The ``\textbf{Linear}" components denote additional linear layers and ``\textbf{Text}" indicates the corresponding text features. Note that ``\textbf{Text}" in this Figure is used to represent the same value. \textbf{\textit{2)}} The orange dashed box represents the pipeline for the few-shot setting. The ``\textbf{memory}" components represent the memory banks. The symbol with a letter C inside a circle denotes the calculation of cosine similarity.}
  \label{fig:main}
\end{figure*}

\subsection{Zero-shot AD Setting}
\noindent
\textbf{Anomaly Classification.} We employ a binary zero-shot anomaly classification framework proposed by~\cite{winclip}, which necessitates descriptions for both normal and abnormal objects. We further improve the detection accuracy by utilizing a text prompt ensemble strategy, including both state and template-level. For the state-level, we adapt generic text to describe the normal and abnormal objects (\eg, flawless, damaged), rather than providing excessively detailed descriptions such as ``chip around edge and corner". For the template-level, we screen the 85 templates in CLIP for ImageNet~\cite{imagenet} and remove some templates that are not suitable for AD, such as ``a photo of the weird [obj.]". 

We combine templates with the states of objects and mean the text features extracted by the text encoder as the final text features $\mathbf{\textit{F}}_t \in \mathbb{R}^{2 \times C}$, where C represents the number of channels of the feature. Let $\mathbf{\textit{F}}_c \in \mathbb{R}^{1 \times C}$ represent the image features used for classification, and the relative probabilities $s$ of the object image being classified as normal or abnormal can be represented as,
\begin{equation}
s = \mathrm{softmax}(\mathbf{\textit{F}}_c {\mathbf{\textit{F}}_t}^T),
\end{equation}
we use the probability corresponding to the anomaly as the anomaly score for the image.

\noindent
\textbf{Anomaly Segmentation.} The structure of the image encoder of CLIP model can be based on either transformer or CNN. Although not used in the classification task, both architectures are capable of extracting features that preserve spacial information. These features can be represented as $\mathbf{\textit{F}}_s^n \in \mathbb{R}^{H \times W \times C_s}$, where $n$ indicates that the features are extracted by the $n$-th stage in the image encoder.

It is a natural idea to obtain the anomaly maps by calculating the similarity between the image features $\mathbf{\textit{F}}_s^n$ and the text features. However, since the CLIP model is designed for classification, image features other than those used for classification are not mapped to the joint embedding space. That is, these features cannot be compared directly to the text features in that space. As a result, we propose to map these image features into the joint embedding space by adding an additional linear layer, and then compare them with the text features.

In addition, we leverage the distinct features offered by the shallow and deep layers of the image encoder. The whole process is represented in the blue dashed box in Fig.~\ref{fig:main}. Taking the transformer based architecture ViT~\cite{vit} as an example, we empirically divide all layers into four stages and use a linear layer for each stage to map the output features into the joint embedding space,
\begin{equation}
{\mathbf{\textit{F}}_s^n}' = k^n \mathbf{\textit{F}}_s^n + b^n,
\end{equation}
where $\mathbf{\textit{F}}_s^n$ here refers to the features of the patch tokens, and $k^n$ and $b^n$ represent weights and bias of the corresponding linear layer respectively. To obtain the anomaly maps for different stages, we make similarity comparison between the adjusted features ${\mathbf{\textit{F}}_s^n}' \in \mathbb{R}^{HW \times C}$ and the text features $\mathbf{\textit{F}}_t$ stage by stage. The final anomaly map $\mathbf{M}$ can be obtained by adding up the results for all the stages,
\begin{equation}
\mathbf{M} = \sum_n \mathrm{softmax}({\mathbf{\textit{F}}_s^n}' {\mathbf{\textit{F}}_t}^T).
\end{equation}
As to the CLIP model based on the CNN architecture, the anomaly maps need to be upsampled to the same resolution before they are added together.

\noindent
\textbf{Losses.} In our approach, the parameters of the CLIP model are frozen, but the linear layers used to map the image features to the joint embedding space require training. Thus, we supervise the anomaly map prediction with the linear combination of the focal loss~\cite{focalloss} and dice loss~\cite{diceloss}. 

\subsection{Few-shot AD Setting}
\label{few_shot}

\noindent
\textbf{Anomaly Classification.} For the few shot setting, the anomaly score of the image comes from two parts. The first part is guided by the text prompt, which is the same as the zero-shot setting. The second part follows the conventional approach used in many AD methods, where the maximum value of the anomaly map is considered. We add the two parts together as the final anomaly score.

\noindent
\textbf{Anomaly Segmentation.} In the few-shot AD setting, only a few (\eg, 1, 5, 10) normal images are available. Memory bank-based approach~\cite{pathcore, winclip} is a promising solution to this problem, making full use of all available normal images in an intuitive way. As shown in the orange dotted line box in Fig.~\ref{fig:main}, we use the image encoder of CLIP to extract the features of the reference images and store them in the feature memory banks $\mathbb{M}^n$, where $n$ indicates that the memory bank stores the output features of stage $n$. In order to make a comprehensive comparison and improve the detection accuracy, we save multi-layer features of the encoder. For the purpose of distinction, we denote the features extracted from the reference images as $\Tilde{F}_s^n$. It is noteworthy that features stored in the memory banks in few-shot come from the exact same stages as the features used to calculate the anomaly maps in zero-shot, unless otherwise noted.

During the test phase, we use the same image encoder to extract the features of different stages of the test image, and then compare them with the reference features in the memory bank of the corresponding stage to get the anomaly maps. The comparison is based on the cosine similarity. Next, we add all the anomaly maps together to get the result $\mathbf{M}_f$. Mathematically,
\begin{equation}
\mathbf{M}_f(i, j) = \sum_n \min_{f \in \mathbb{M}^n}(1 - \langle {\mathbf{\textit{F}}_s^n(i,j)}, f \rangle).
\end{equation}
Finally, the anomaly map $\mathbf{M}_f$ obtained by feature comparison is added with the anomaly map $\mathbf{M}$ obtained by zero-shot as the final result. It is worth noting that we do not fine-tune the linear layers with the reference images in the few-shot setting, but instead directly utilize the weights obtained in the zero-shot setting.

\begin{figure*}
  \centering
  \includegraphics[width=\linewidth]{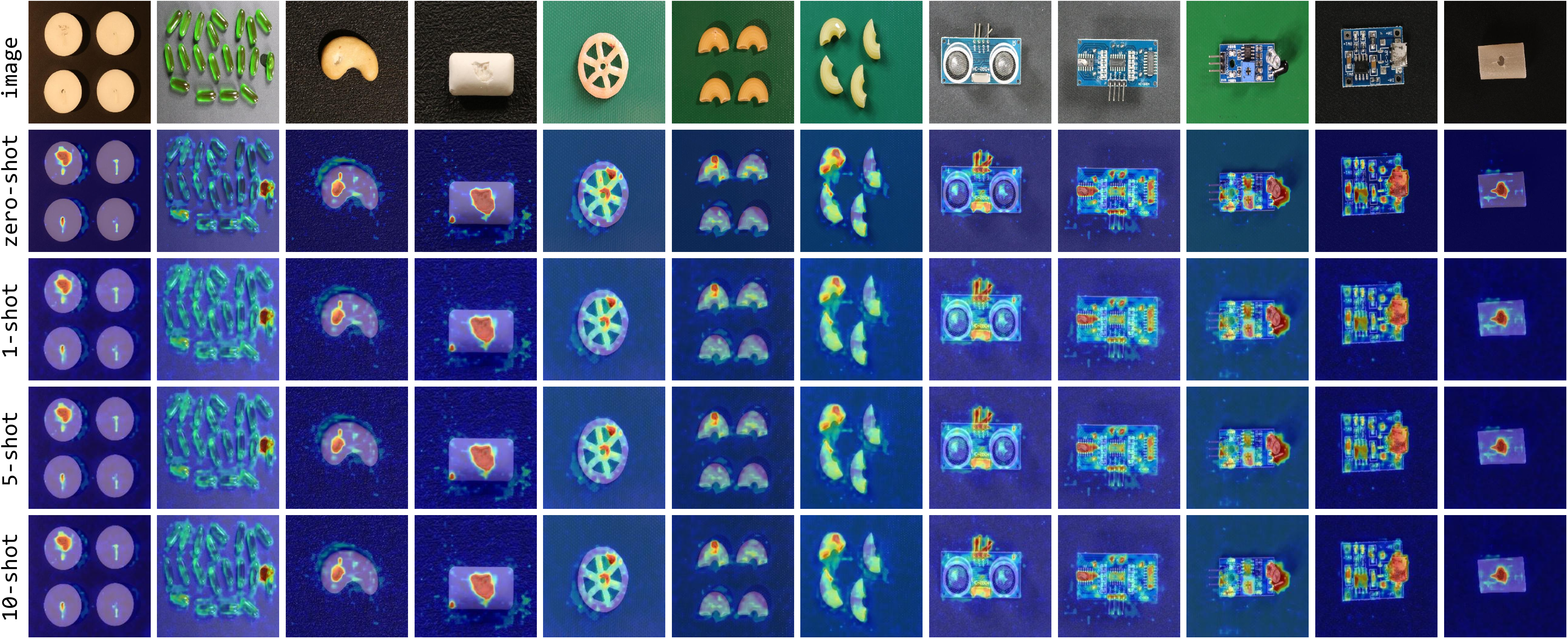}
  \caption{\textbf{Results visualizations on zero-/few-shot settings}. The first row shows the original image, the second row displays the zero-shot results, and the third to fifth rows present the results for 1-shot, 5-shot, and 10-shot, respectively.}
  \label{fig:visualization}
\end{figure*}

\section{Experiments}
\subsection{Setup}
\noindent
\textbf{Datasets.} The organizers of the VAND Challenge utilize a modified VisA~\cite{visa} dataset to evaluate the effectiveness of different models, and we refer to this dataset as the VAND test set. \textbf{\textit{1)}} In the zero-shot track, according to the challenge regulations, models can be trained and pre-trained on any permitted datasets, except for the train and test sets of VisA~\cite{visa}. Moreover, as the training of the newly introduced linear layers relies on the ground-truth anomaly maps, \textbf{\textit{we solely use the test set of the MVTec AD~\cite{mvtecad} dataset}}. \textbf{\textit{2)}} In the few-shot track, we incorporate the linear layers without any additional fine-tuning using the reference images, as discussed in Sec.~\ref{few_shot}.

To thoroughly demonstrate the effectiveness of our approach, we also conducted experiments on the standard MVTec AD~\cite{mvtecad} and VisA~\cite{visa} dataset. For the zero-shot setting, when testing on the MVTec AD, we trained the linear layers using the test set of VisA. Conversely, when testing on VisA, we trained the linear layers using the test set of MVTec AD. For the few-shot setting, we did not perform any extra fine-tuning on the linear layers either.

\noindent
\textbf{Metrics.} In VAND Challenge, for both classification and segmentation, the evaluation metric is the F1-max score, \ie, F1 score at the optimal threshold. For each category, the harmonic mean between the classification F1-max and segmentation F1-max are chosen as the summary metric. The final metric to compare different models is the arithmetic mean over all categories.
For the standard MVTec AD and VisA dataset, we additionally employ the Area Under the Receiver Operating Curve (AUROC), Average Precision (AP) and per-region-overlap (PRO) as the evaluation metrics. The AUROC and AP metrics can be used for both classification and segmentation, while PRO is specifically applicable to the segmentation evaluation.

\noindent
\textbf{Backbone for feature extraction.} By default, we use the CLIP model with ViT-L/14 and the image resolution is 336. It consists of a total of 24 layers, which we arbitrarily divide into 4 stages, with each stage containing 6 layers. Therefore, we require four additional linear layers to map the image features from the four stages to the joint embedding space. In the few-shot track, we separately store the features of reference images from these four stages.

\begin{table}
  \centering
  \caption{Quantitative results of the top five participating teams on the \textbf{zero-shot} track leaderboard of the VAND 2023 Challenge.}
  \label{tab:zero_shot}
  \tablestyle{6pt}{1.05}
  \setlength\tabcolsep{1.0pt}
  \begin{tabular}{@{}l p{1cm}<{\centering} p{1.6cm}<{\centering} p{1.4cm}<{\centering} p{0.7cm}<{\centering}@{}}
    \toprule
    Team Name & F1-max & F1-max-segm & F1-max-cls & Rank\\
    \midrule
    AaxJIjQ & 0.2788 & 0.2019 & 0.7742 & 5\\
    MediaBrain & 0.2880 & 0.1866 & \textbf{0.7945} & 4 \\
    Variance Vigilance Vanguard & 0.3217 & 0.2197 & \underline{0.7928} & 3\\
    SegmentAnyAnomaly & \underline{0.3956} & \underline{0.2942} & 0.7517 & 2\\
    \midrule
    APRIL-GAN (Ours) & \textbf{0.4589} & \textbf{0.3431} & 0.7782 & 1\\
    \bottomrule
  \end{tabular}
\end{table}

\noindent
\textbf{Training.} We utilize a resolution of 518 $\times$ 518 for the images, incorporating a proposed data augmentation approach. Specifically, we concatenate four images of the same category from the MVTec AD dataset with a 20\% probability to create a new composite image. This is because the VisA dataset contains many instances where a single image includes multiple objects of the same category, which is not present in the MVTec AD. We employ the Adam optimizer with a fixed learning rate of 1e$^{-3}$. Our training process is highly efficient. To enable the model to recognize both normal and abnormal objects while preventing overfitting, we only need to train for 3 epochs with a batch size of 16 on a single GPU (NVIDIA GeForce RTX 3090).
For the standard VisA dataset, we employ the same training strategies. However, for the MVTec AD dataset, the training epochs is set to 15, while keeping the other strategies unchanged.

\begin{table}
  \centering
  \caption{Quantitative results of the top five participating teams on the \textbf{few-shot} track leaderboard of the VAND 2023 Challenge.}
  \label{tab:few_shot}
  \tablestyle{6pt}{1.05}
  \setlength\tabcolsep{1.0pt}
  \begin{tabular}{@{}l p{1cm}<{\centering} p{1.6cm}<{\centering} p{1.4cm}<{\centering} p{0.7cm}<{\centering}@{}}
    \toprule
    Team Name & F1-max & F1-max-segm & F1-max-cls & Rank\\
    \midrule
    VAND-Organizer (WinCLIP) & 0.5323 & 0.4118 & 0.8114 & 5\\
    PatchCore+ & 0.5742 & \underline{0.4542} & 0.8423 & 3\\
    MediaBrain & \underline{0.5763} & 0.4515 & \underline{0.8480} & 2 \\
    Scortex & \textbf{0.5909} & \textbf{0.4706} & 0.8399 & 1\\
    \midrule
    APRIL-GAN (Ours) & 0.5629 & 0.4264 & \textbf{0.8687} & 4\\
    \bottomrule
  \end{tabular}
\end{table}

\subsection{Qualitative results for VAND}
We provide qualitative results for both the zero-/few-shot tracks. As shown in the second row of Fig.~\ref{fig:visualization}, in the zero-shot setting, thanks to the newly introduced linear layers, our method is capable of detecting anomalies with relatively high accuracy. Particularly for categories such as the candle, chewing gum, and pipe fryum, our model not only accurately localizes anomalies, but also has minimal false detections in the background. However, when it comes to more complex categories like macaroni and PCBs, our model, while able to pinpoint the exact locations of anomalies, sometimes misclassifies certain normal regions as anomalies. For instance, some normal components on the PCBs are also identified as anomalies. This could be inevitable since, in the zero-shot setting, the model lacks prior knowledge of the specific components and their expected positions, making it challenging to accurately determine what should be considered abnormal. Reference images are necessary in these cases.

\begin{table*}
  \centering
  \caption{Quantitative comparisons on the \textbf{MVTec AD}~\cite{mvtecad} dataset. We report the mean and standard deviation over 5 random seeds for each measurement. Bold indicates the best performance, while underline denotes the second-best result.}
  \label{tab:comp_mvtec}
  \tablestyle{6pt}{1.05}
  \setlength\tabcolsep{1.0pt}
  \begin{tabular}{@{}l p{2cm}<{\centering} p{1.8cm}<{\centering} p{1.8cm}<{\centering} p{1.8cm}<{\centering} p{1.8cm}<{\centering} p{1.8cm}<{\centering} p{1.8cm}<{\centering} p{1.8cm}<{\centering}@{}}
    \toprule
    Setting & Method & AUROC-segm & F1-max-segm & AP-segm & PRO-segm & AUROC-cls & F1-max-cls & AP-cls\\
    \midrule
    \multirow{2}{*}{zero-shot} & WinCLIP~\cite{winclip} & \underline{85.1} & \underline{31.7} & - & \textbf{64.6} & \textbf{91.8} & \textbf{92.9} & \textbf{96.5} \\
    ~ & Ours & \textbf{87.6} & \textbf{43.3} & \textbf{40.8} & \underline{44.0} & \underline{86.1} & \underline{90.4} & \underline{93.5} \\
    \midrule
    \multirow{5}{*}{1-shot} & SPADE~\cite{spade} & 92.0\scriptsize{$\pm$0.3} & 44.5\scriptsize{$\pm$1.0} & - & 85.7\scriptsize{$\pm$0.7} & 82.9\scriptsize{$\pm$2.6} & 91.1\scriptsize{$\pm$1.0} & 91.7\scriptsize{$\pm$1.2}  \\
    ~ & PaDiM~\cite{padim} & 91.3\scriptsize{$\pm$0.7} & 43.7\scriptsize{$\pm$1.5} & - & 78.2\scriptsize{$\pm$1.8} & 78.9\scriptsize{$\pm$3.1} & 89.2\scriptsize{$\pm$1.1} & 89.3\scriptsize{$\pm$1.7} \\
    ~ & PatchCore~\cite{pathcore} & 93.3\scriptsize{$\pm$0.6} & 53.0\scriptsize{$\pm$1.7} & - & 82.3\scriptsize{$\pm$1.3} & 86.3\scriptsize{$\pm$3.3} & 92.0\scriptsize{$\pm$1.5} & 93.8\scriptsize{$\pm$1.7} \\
    ~ & WinCLIP~\cite{winclip} & \textbf{95.2\scriptsize{$\pm$0.5}} & \textbf{55.9\scriptsize{$\pm$2.7}} & - & \underline{87.1\scriptsize{$\pm$1.2}} & \textbf{93.1\scriptsize{$\pm$2.0}} & \textbf{93.7\scriptsize{$\pm$1.1}} & \textbf{96.5\scriptsize{$\pm$0.9}} \\
    ~ & Ours & \underline{95.1\scriptsize{$\pm$0.1}} & \underline{54.2\scriptsize{$\pm$0.0}} & \textbf{51.8\scriptsize{$\pm$0.1}} & \textbf{90.6\scriptsize{$\pm$0.2}} & \underline{92.0\scriptsize{$\pm$0.3}} & \underline{92.4\scriptsize{$\pm$0.2}} & \underline{95.8\scriptsize{$\pm$0.2}} \\
    \midrule
    \multirow{5}{*}{2-shot} & SPADE~\cite{spade} & 91.2\scriptsize{$\pm$0.4} & 42.4\scriptsize{$\pm$1.0} & - & 83.9\scriptsize{$\pm$0.7} & 81.0\scriptsize{$\pm$2.0} & 90.3\scriptsize{$\pm$0.8} & 90.6\scriptsize{$\pm$0.8}  \\
    ~ & PaDiM~\cite{padim} & 89.3\scriptsize{$\pm$0.9} & 40.2\scriptsize{$\pm$2.1} & - & 73.3\scriptsize{$\pm$2.0} & 76.6\scriptsize{$\pm$3.1} & 88.2\scriptsize{$\pm$1.1} & 88.1\scriptsize{$\pm$1.7} \\
    ~ & PatchCore~\cite{pathcore} & 92.0\scriptsize{$\pm$1.0} & 50.4\scriptsize{$\pm$2.1} & - & 79.7\scriptsize{$\pm$2.0} & 83.4\scriptsize{$\pm$3.0} & 90.5\scriptsize{$\pm$1.5} & 92.2\scriptsize{$\pm$1.5} \\
    ~ & WinCLIP~\cite{winclip} & \textbf{96.0\scriptsize{$\pm$0.3}} & \textbf{58.4\scriptsize{$\pm$1.7}} & - & \underline{88.4\scriptsize{$\pm$0.9}} & \textbf{94.4\scriptsize{$\pm$1.3}} & \textbf{94.4\scriptsize{$\pm$0.8}} & \textbf{97.0\scriptsize{$\pm$0.7}} \\
    ~ & Ours & \underline{95.5\scriptsize{$\pm$0.0}} & \underline{55.9\scriptsize{$\pm$0.5}} & \textbf{53.4\scriptsize{$\pm$0.4}} & \textbf{91.3\scriptsize{$\pm$0.1}} & \underline{92.4\scriptsize{$\pm$0.3}} & \underline{92.6\scriptsize{$\pm$0.1}} & \underline{96.0\scriptsize{$\pm$0.2}} \\
    \midrule
    \multirow{5}{*}{4-shot} & SPADE~\cite{spade} & 92.7\scriptsize{$\pm$0.3} & 46.2\scriptsize{$\pm$1.3} & - & 87.0\scriptsize{$\pm$0.5} & 84.8\scriptsize{$\pm$2.5} & 91.5\scriptsize{$\pm$0.9} & 92.5\scriptsize{$\pm$1.2}  \\
    ~ & PaDiM~\cite{padim} & 92.6\scriptsize{$\pm$0.7} & 46.1\scriptsize{$\pm$1.8} & - & 81.3\scriptsize{$\pm$1.9} & 80.4\scriptsize{$\pm$2.5} & 90.2\scriptsize{$\pm$1.2} & 90.5\scriptsize{$\pm$1.6} \\
    ~ & PatchCore~\cite{pathcore} & 94.3\scriptsize{$\pm$0.5} & 55.0\scriptsize{$\pm$1.9} & - & 84.3\scriptsize{$\pm$1.6} & 88.8\scriptsize{$\pm$2.6} & 92.6\scriptsize{$\pm$1.6} & 94.5\scriptsize{$\pm$1.5} \\
    ~ & WinCLIP~\cite{winclip} & \textbf{96.2\scriptsize{$\pm$0.3}} & \textbf{59.5\scriptsize{$\pm$1.8}} & - & \underline{89.0\scriptsize{$\pm$0.8}} & \textbf{95.2\scriptsize{$\pm$1.3}} & \textbf{94.7\scriptsize{$\pm$0.8}} & \textbf{97.3\scriptsize{$\pm$0.6}} \\
    ~ & Ours & \underline{95.9\scriptsize{$\pm$0.0}} & \underline{56.9\scriptsize{$\pm$0.1}} & \textbf{54.5\scriptsize{$\pm$0.2}} & \textbf{91.8\scriptsize{$\pm$0.1}} & \underline{92.8\scriptsize{$\pm$0.2}} & \underline{92.8\scriptsize{$\pm$0.1}} & \underline{96.3\scriptsize{$\pm$0.1}} \\
    \bottomrule
  \end{tabular}
\end{table*}

\begin{table*}
  \centering
  \caption{Quantitative comparisons on the \textbf{VisA}~\cite{visa} dataset. We report the mean and standard deviation over 5 random seeds for each measurement. Bold indicates the best performance, while underline denotes the second-best result.}
  \label{tab:comp_visa}
  \tablestyle{6pt}{1.05}
  \setlength\tabcolsep{1.0pt}
  \begin{tabular}{@{}l p{2cm}<{\centering} p{1.8cm}<{\centering} p{1.8cm}<{\centering} p{1.8cm}<{\centering} p{1.8cm}<{\centering} p{1.8cm}<{\centering} p{1.8cm}<{\centering} p{1.8cm}<{\centering}@{}}
    \toprule
    Setting & Method & AUROC-segm & F1-max-segm & AP-segm & PRO-segm & AUROC-cls & F1-max-cls & AP-cls\\
    \midrule
    \multirow{2}{*}{zero-shot} & WinCLIP~\cite{winclip} & \underline{79.6} & \underline{14.8} & - & \underline{56.8} & \textbf{78.1} & \textbf{79.0} & \underline{81.2} \\
    ~ & Ours & \textbf{94.2} & \textbf{32.3} & \textbf{25.7} & \textbf{86.8} & \underline{78.0} & \underline{78.7} & \textbf{81.4} \\
    \midrule
    \multirow{5}{*}{1-shot} & SPADE~\cite{spade} & 95.6\scriptsize{$\pm$0.4} & 35.5\scriptsize{$\pm$2.2} & - & 84.1\scriptsize{$\pm$1.6} & 79.5\scriptsize{$\pm$4.0} & 80.7\scriptsize{$\pm$1.9} & 82.0\scriptsize{$\pm$3.3}  \\
    ~ & PaDiM~\cite{padim} & 89.9\scriptsize{$\pm$0.8} & 17.4\scriptsize{$\pm$1.7} & - & 64.3\scriptsize{$\pm$2.4} & 62.8\scriptsize{$\pm$5.4} & 75.3\scriptsize{$\pm$1.2} & 68.3\scriptsize{$\pm$4.0} \\
    ~ & PatchCore~\cite{pathcore} & 95.4\scriptsize{$\pm$0.6} & 38.0\scriptsize{$\pm$1.9} & - & 80.5\scriptsize{$\pm$2.5} & 79.9\scriptsize{$\pm$2.9} & 81.7\scriptsize{$\pm$1.6} & 82.8\scriptsize{$\pm$2.3} \\
    ~ & WinCLIP~\cite{winclip} & \textbf{96.4\scriptsize{$\pm$0.4}} & \textbf{41.3\scriptsize{$\pm$2.3}} & - & \underline{85.1\scriptsize{$\pm$2.1}} & \underline{83.8\scriptsize{$\pm$4.0}} & \underline{83.1\scriptsize{$\pm$1.7}} & \underline{85.1\scriptsize{$\pm$4.0}} \\ 
    ~ & Ours & \underline{96.0\scriptsize{$\pm$0.0}} & \underline{38.5\scriptsize{$\pm$0.3}} & \textbf{30.9\scriptsize{$\pm$0.3}} & \textbf{90.0\scriptsize{$\pm$0.1}} & \textbf{91.2\scriptsize{$\pm$0.8}} & \textbf{86.9\scriptsize{$\pm$0.6}} & \textbf{93.3\scriptsize{$\pm$0.8}} \\ 
    \midrule
    \multirow{5}{*}{2-shot} & SPADE~\cite{spade} & \underline{96.2\scriptsize{$\pm$0.4}} & 40.5\scriptsize{$\pm$3.7} & - & 85.7\scriptsize{$\pm$1.1} & 80.7\scriptsize{$\pm$5.0} & 81.7\scriptsize{$\pm$2.5} & 82.3\scriptsize{$\pm$4.3}  \\
    ~ & PaDiM~\cite{padim} & 92.0\scriptsize{$\pm$0.7} & 21.1\scriptsize{$\pm$2.4} & - & 70.1\scriptsize{$\pm$2.6} & 67.4\scriptsize{$\pm$5.1} & 75.7\scriptsize{$\pm$1.8} & 71.6\scriptsize{$\pm$3.8} \\
    ~ & PatchCore~\cite{pathcore} & 96.1\scriptsize{$\pm$0.5} & \underline{41.0\scriptsize{$\pm$3.9}} & - & 82.6\scriptsize{$\pm$2.3} & 81.6\scriptsize{$\pm$4.0} & 82.5\scriptsize{$\pm$1.8} & 84.8\scriptsize{$\pm$3.2} \\
    ~ & WinCLIP~\cite{winclip} & \textbf{96.8\scriptsize{$\pm$0.3}} & \textbf{43.5\scriptsize{$\pm$3.3}} & - & \underline{86.2\scriptsize{$\pm$1.4}} & \underline{84.6\scriptsize{$\pm$2.4}} & \underline{83.0\scriptsize{$\pm$1.4}} & \underline{85.8\scriptsize{$\pm$2.7}} \\
    ~ & Ours & \underline{96.2\scriptsize{$\pm$0.0}} & 39.3\scriptsize{$\pm$0.2} & \textbf{31.6\scriptsize{$\pm$0.3}} & \textbf{90.1\scriptsize{$\pm$0.1}} & \textbf{92.2\scriptsize{$\pm$0.3}} & \textbf{87.7\scriptsize{$\pm$0.3}} & \textbf{94.2\scriptsize{$\pm$0.3}} \\
    \midrule
    \multirow{5}{*}{4-shot} & SPADE~\cite{spade} & 96.6\scriptsize{$\pm$0.3} & 43.6\scriptsize{$\pm$3.6} & - & 87.3\scriptsize{$\pm$0.8} & 81.7\scriptsize{$\pm$3.4} & 82.1\scriptsize{$\pm$2.1} & 83.4\scriptsize{$\pm$2.7}  \\
    ~ & PaDiM~\cite{padim} & 93.2\scriptsize{$\pm$0.5} & 24.6\scriptsize{$\pm$1.8} & - & 72.6\scriptsize{$\pm$1.9} & 72.8\scriptsize{$\pm$2.9} & 78.0\scriptsize{$\pm$1.2} & 75.6\scriptsize{$\pm$2.2} \\
    ~ & PatchCore~\cite{pathcore} & 96.8\scriptsize{$\pm$0.3} & \underline{43.9\scriptsize{$\pm$3.1}} & - & 84.9\scriptsize{$\pm$1.4} & 85.3\scriptsize{$\pm$2.1} & 84.3\scriptsize{$\pm$1.3} & 87.5\scriptsize{$\pm$2.1} \\
    ~ & WinCLIP~\cite{winclip} & \textbf{97.2\scriptsize{$\pm$0.2}} & \textbf{47.0\scriptsize{$\pm$3.0}} & - & \underline{87.6\scriptsize{$\pm$0.9}} & \underline{87.3\scriptsize{$\pm$1.8}} & \underline{84.2\scriptsize{$\pm$1.6}} & \underline{88.8\scriptsize{$\pm$1.8}} \\
    ~ & Ours & \underline{96.2\scriptsize{$\pm$0.0}} & 40.0\scriptsize{$\pm$0.1} & \textbf{32.2\scriptsize{$\pm$0.1}} & \textbf{90.2\scriptsize{$\pm$0.1}} & \textbf{92.6\scriptsize{$\pm$0.4}} & \textbf{88.4\scriptsize{$\pm$0.5}} & \textbf{94.5\scriptsize{$\pm$0.3}} \\
    \bottomrule
  \end{tabular}
\end{table*}

As illustrated in the last three rows of Fig.~\ref{fig:visualization}, for simple categories, our method can effectively maintain the performance achieved in the zero-shot setting and achieve precise anomaly localization in the few-shot setting. For complex categories, our method, when combined with reference images, significantly reduces false detection rates. For example, in the case of macaroni in the 7th column, the extent of the anomalous region is narrowed down. Similarly, for PCB in the 10th column, the anomaly scores for normal components are greatly reduced.

\subsection{Quantitative results for VAND}
Due to the unavailability of the ground-truth anomaly maps for the VAND test set, we present the top 5 results from the leaderboard for both tracks. \\
\noindent\textbf{Zero-shot setting.} As shown in the Tab.~\ref{tab:zero_shot}, our method achieved F1 scores of 0.3431 for anomaly segmentation and 0.7782 for anomaly classification in the zero-shot track, ranking first overall. Specifically, our method ranked first in the anomaly segmentation and fifth in the anomaly classification individually. Notebaly, our method significantly outperforms the second-ranked team in terms of segmentation, with a large margin of 0.0489.\\
\noindent\textbf{Few-shot setting.} As to the few-shot track, although we only ranked fourth in the Tab.~\ref{tab:few_shot}, it is notable that we ranked first in the anomaly classification with an F1 score of 0.8687. This score also considerably surpasses the second-ranked team, exceeding them by 0.0207.

\begin{figure}[tp]
    \centering
    \includegraphics[width=0.95\linewidth]{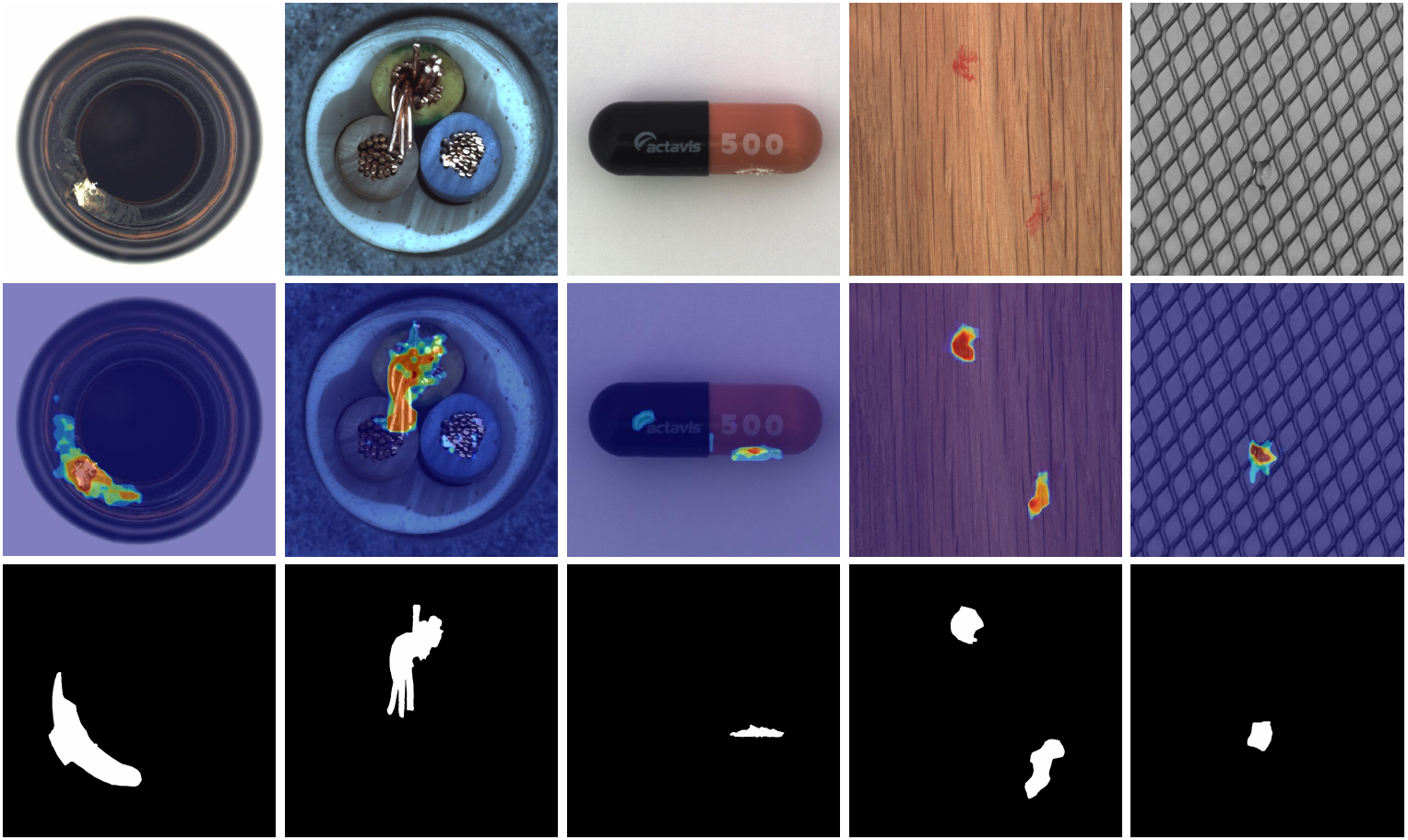}
    \caption{Visualization results on the \textbf{MVTec AD}~\cite{mvtecad} dataset under the \textbf{zero-shot} setting.}
    \label{fig:mvtec}
\end{figure}

\subsection{Quantitative and qualitative results for MVTec}
\label{sec:mvtec}
Tab.~\ref{tab:comp_mvtec} presents the comparative results of our method and other approaches on the MVTec AD dataset. In the zero-shot setting, our method outperforms WinCLIP~\cite{winclip} by a large margin in terms of AUROC-segm (2.5$\uparrow$) and F1-max-segm (11.6$\uparrow$) metrics, but it achieves a lower PRO metric. We believe that this phenomenon may be related to the definition of the PRO metric. PRO essentially calculates the proportion of pixels successfully detected as anomalies within each connected component of anomalies. When the identified anomaly regions can completely cover the ground truth, or even exceed it, the PRO metric tends to be high. However, if the identified anomaly region is relatively small, even if its position is correct, the PRO metric may be lower. As shown in~\cref{fig:mvtec}, our method tends to identify a precise anomaly region, which is often slightly smaller than the ground truth, despite being in the correct location. This could be attributed to training the linear layers on the VisA dataset, as compared to the MVTec AD dataset, anomalies in VisA tend to be smaller in size. As to anomaly classification, our method slightly lags behind WinCLIP.

In the few-shot setting, our method achieves slightly lower results compared to WinCLIP on most metrics, but significantly outperforms other comparison methods. Furthermore, in contrast to the zero-shot setting, our PRO metric surpasses that of WinCLIP, which can be attributed to the design of our multiple memory banks.

Additionally, the detailed results for each category in the MVTec AD dataset under the zero-shot, 1-shot, 2-shot, and 4-shot settings are recorded in~\cref{tab:zero_shot_mvtec},~\cref{tab:1_shot_mvtec},~\cref{tab:2_shot_mvtec} and~\cref{tab:4_shot_mvtec}, respectively.

\begin{figure}[tp]
    \centering
    \includegraphics[width=0.95\linewidth]{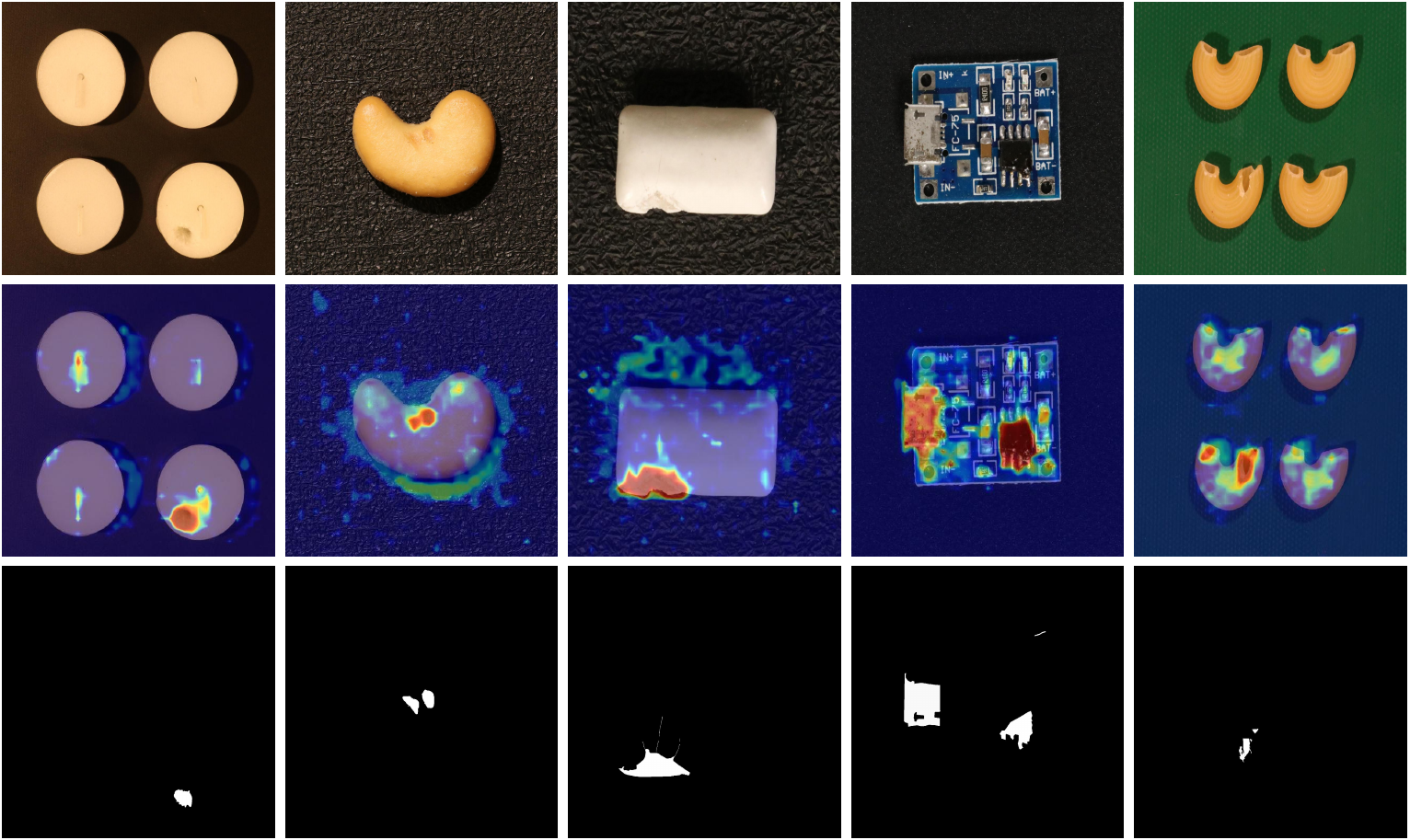}
    \caption{Visualization results on the \textbf{VisA}~\cite{visa} dataset under the \textbf{zero-shot} setting.}
    \label{fig:visa}
\end{figure}

\subsection{Quantitative and qualitative results for VisA}
Tab.~\ref{tab:comp_visa} presents the comparative results of our method and other approaches on the VisA dataset. In the zero-shot setting, our method significantly outperforms WinCLIP~\cite{winclip} in metrics AUROC-segm (14.6$\uparrow$), F1-max-segm (17.5$\uparrow$), and PRO-segm (30$\uparrow$). The visualization results are presented in~\cref{fig:visa}. In this section, we trained on the MVTec AD dataset and tested on the VisA dataset. As the anomalies in MVTec AD tend to be larger overall, the detected anomaly regions on VisA also tend to be slightly larger than the ground truth, resulting in higher PRO scores. This is consistent with the discussion in~\cref{sec:mvtec}. Although it slightly lags behind in anomaly classification, the difference is minimal, and in fact, our method even achieves a slightly higher AP metric compared to WinCLIP. 

In the few-shot setting, our method falls slightly behind WinCLIP in terms of anomaly segmentation, but achieves a higher PRO metric. It is worth noting that our method significantly outperforms WinCLIP and other methods in all metrics related to anomaly classification.

Additionally, the detailed results for each category in the VisA dataset under the zero-shot, 1-shot, 2-shot, and 4-shot settings are recorded in~\cref{tab:zero_shot_visa},~\cref{tab:1_shot_visa},~\cref{tab:2_shot_visa} and~\cref{tab:4_shot_visa}, respectively.

\section{Conclusion}
In reality, due to the diverse range of industrial products, it is not feasible to collect a large number of training images or train specialized models for each category. Thus, designing joint models for zero/few-shot settings is a promising research direction. We made slight modifications to the CLIP model by adding extra linear layers, enabling it to perform zero-shot segmentation while still being capable of zero-shot classification. Moreover, we further improved the performance by incorporating memory banks along with a few reference images. Our approach achieved first place in the zero-shot track and fourth place in the few-shot track of the VAND Challenge.\\
\noindent\textbf{Acknowledgments.} This work is support by the Youtu Lab, Tencent, and we appreciate the constructive advice and guidance provided by Yabiao Wang (Researcher in Youtu Lab, Tencent), Chengjie Wang (Researcher in Youtu Lab, Tencent), and Yong Liu (Prof. in APRIL Lab, Zhejiang University).

\begin{table*}
  \centering
  \caption{Quantitative results of the \textbf{zero-shot} setting on the \textbf{MVTec AD}~\cite{mvtecad} dataset.}
  \label{tab:zero_shot_mvtec}
  \tablestyle{6pt}{1.05}
  \setlength\tabcolsep{1.0pt}
  \begin{tabular}{@{}l p{1.1cm}<{\centering} p{1.1cm}<{\centering} p{1.1cm}<{\centering} p{1.1cm}<{\centering} p{1.1cm}<{\centering} p{1.1cm}<{\centering} p{1.1cm}<{\centering}@{}}
    \toprule
    Object & AUROC-segm & F1-max-segm & AP-segm & PRO-segm & AUROC-cls & F1-max-cls & AP-cls\\
    \midrule
    carpet & 98.4 & 65.7 & 67.5 & 48.5 & 99.5 & 98.3 & 99.8\\
    bottle&83.4&53.4&53.0&45.6&92.0&92.8&97.7\\
    hazelnut&96.1&50.5&49.7&70.3&89.6&87.7&94.8\\
    leather&99.1&50.0&52.3&72.4&99.7&98.9&99.9\\
    cable&72.3&23.9&18.2&25.7&88.4&84.6&93.1\\
    capsule&92.0&33.1&29.7&51.3&79.9&91.6&95.5\\
    grid&95.8&40.7&36.6&31.6&86.3&89.1&94.9\\
    pill&76.2&27.7&23.6&65.4&80.5&91.6&96.0\\
    transistor&62.4&19.0&11.7&21.3&80.8&73.1&77.5\\
    metal\_nut&65.4&28.1&25.9&38.4&68.4&89.4&91.9\\
    screw&97.8&41.7&33.7&67.1&84.9&89.3&93.6\\
    toothbrush&95.8&48.1&43.2&54.5&53.8&83.3&71.5\\
    zipper&91.1&40.5&38.7&10.7&89.6&90.8&97.1\\
    tile&92.7&66.5&66.3&26.7&99.9&99.4&100.0\\
    wood&95.8&60.3&61.8&31.1&99.0&96.8&99.7\\
    \midrule
    Mean&87.6&43.3&40.8&44.0&86.1&90.4&93.5\\
    \bottomrule
  \end{tabular}
\end{table*}

\begin{table*}
  \centering
  \caption{Quantitative results of the \textbf{1-shot} setting on the \textbf{MVTec AD}~\cite{mvtecad} dataset. We report the mean and standard deviation over 5 random seeds for each measurement.}
  \label{tab:1_shot_mvtec}
  \tablestyle{6pt}{1.05}
  \setlength\tabcolsep{1.0pt}
  \begin{tabular}{@{}l p{1.8cm}<{\centering} p{1.8cm}<{\centering} p{1.8cm}<{\centering} p{1.8cm}<{\centering} p{1.8cm}<{\centering} p{1.8cm}<{\centering} p{1.8cm}<{\centering}@{}}
    \toprule
    Object & AUROC-segm & F1-max-segm & AP-segm & PRO-segm & AUROC-cls & F1-max-cls & AP-cls\\
    \midrule
    carpet & 98.7\scriptsize{$\pm$0.0} & 71.6\scriptsize{$\pm$0.2} & 71.6\scriptsize{$\pm$0.1} & 96.5\scriptsize{$\pm$0.2} & 99.9\scriptsize{$\pm$0.0} & 99.0\scriptsize{$\pm$0.2} & 100.0\scriptsize{$\pm$0.0} \\
    bottle & 96.4\scriptsize{$\pm$0.1} & 71.6\scriptsize{$\pm$0.8} & 73.7\scriptsize{$\pm$0.7} & 92.7\scriptsize{$\pm$0.1} & 92.8\scriptsize{$\pm$0.3} & 92.3\scriptsize{$\pm$1.1} & 97.8\scriptsize{$\pm$0.1} \\
    hazelnut & 97.6\scriptsize{$\pm$0.1} & 59.2\scriptsize{$\pm$1.2} & 59.3\scriptsize{$\pm$1.2} & 92.1\scriptsize{$\pm$0.3} & 98.5\scriptsize{$\pm$0.1} & 96.8\scriptsize{$\pm$0.3} & 99.3\scriptsize{$\pm$0.0} \\
    leather & 99.5\scriptsize{$\pm$0.0} & 51.6\scriptsize{$\pm$0.3} & 55.9\scriptsize{$\pm$0.1} & 99.0\scriptsize{$\pm$0.1} & 99.9\scriptsize{$\pm$0.1} & 99.6\scriptsize{$\pm$0.2} & 100.0\scriptsize{$\pm$0.0} \\
    cable & 90.8\scriptsize{$\pm$0.5} & 36.3\scriptsize{$\pm$1.4} & 31.6\scriptsize{$\pm$0.8} & 81.5\scriptsize{$\pm$1.3} & 74.7\scriptsize{$\pm$1.2} & 78.2\scriptsize{$\pm$0.4} & 84.8\scriptsize{$\pm$0.9} \\
    capsule & 97.1\scriptsize{$\pm$0.3} & 37.0\scriptsize{$\pm$4.2} & 36.3\scriptsize{$\pm$3.8} & 95.6\scriptsize{$\pm$0.7} & 92.0\scriptsize{$\pm$2.2} & 92.8\scriptsize{$\pm$0.4} & 98.3\scriptsize{$\pm$0.6} \\
    grid & 96.9\scriptsize{$\pm$0.3} & 47.7\scriptsize{$\pm$1.2} & 41.8\scriptsize{$\pm$0.9} & 90.2\scriptsize{$\pm$0.7} & 99.0\scriptsize{$\pm$0.3} & 98.1\scriptsize{$\pm$0.8} & 99.6\scriptsize{$\pm$0.1} \\
    pill & 94.8\scriptsize{$\pm$0.5} & 54.4\scriptsize{$\pm$1.1} & 46.4\scriptsize{$\pm$1.3} & 96.6\scriptsize{$\pm$0.2} & 84.6\scriptsize{$\pm$0.7} & 92.4\scriptsize{$\pm$0.2} & 97.0\scriptsize{$\pm$0.2} \\
    transistor & 80.4\scriptsize{$\pm$1.4} & 35.5\scriptsize{$\pm$2.6} & 25.5\scriptsize{$\pm$1.9} & 64.8\scriptsize{$\pm$1.4} & 83.4\scriptsize{$\pm$1.2} & 75.0\scriptsize{$\pm$1.5} & 76.1\scriptsize{$\pm$2.6} \\
    metal\_nut & 89.6\scriptsize{$\pm$0.9} & 54.1\scriptsize{$\pm$2.0} & 53.1\scriptsize{$\pm$2.0} & 88.2\scriptsize{$\pm$0.8} & 88.5\scriptsize{$\pm$1.2} & 90.6\scriptsize{$\pm$0.6} & 97.4\scriptsize{$\pm$0.3} \\
    screw & 98.2\scriptsize{$\pm$0.0} & 42.1\scriptsize{$\pm$0.7} & 34.4\scriptsize{$\pm$0.5} & 92.6\scriptsize{$\pm$0.2} & 79.8\scriptsize{$\pm$0.6} & 88.4\scriptsize{$\pm$0.2} & 91.1\scriptsize{$\pm$0.4} \\
    toothbrush & 98.4\scriptsize{$\pm$0.3} & 55.3\scriptsize{$\pm$1.2} & 55.1\scriptsize{$\pm$1.8} & 93.4\scriptsize{$\pm$1.1} & 94.3\scriptsize{$\pm$2.1} & 93.0\scriptsize{$\pm$1.5} & 97.8\scriptsize{$\pm$0.8} \\
    zipper & 96.1\scriptsize{$\pm$0.2} & 56.3\scriptsize{$\pm$0.9} & 54.6\scriptsize{$\pm$1.0} & 87.2\scriptsize{$\pm$0.7} & 94.3\scriptsize{$\pm$0.3} & 94.6\scriptsize{$\pm$0.7} & 98.3\scriptsize{$\pm$0.1} \\
    tile & 95.4\scriptsize{$\pm$0.2} & 72.0\scriptsize{$\pm$0.5} & 69.4\scriptsize{$\pm$0.5} & 92.3\scriptsize{$\pm$0.3} & 98.9\scriptsize{$\pm$0.1} & 97.4\scriptsize{$\pm$0.3} & 99.6\scriptsize{$\pm$0.0} \\
    wood & 96.1\scriptsize{$\pm$0.1} & 67.9\scriptsize{$\pm$0.3} & 69.0\scriptsize{$\pm$0.5} & 95.9\scriptsize{$\pm$0.1} & 98.5\scriptsize{$\pm$0.1} & 97.5\scriptsize{$\pm$0.0} & 99.5\scriptsize{$\pm$0.0} \\
    \midrule
    Mean & 95.1\scriptsize{$\pm$0.1} & 54.2\scriptsize{$\pm$0.0} & 51.8\scriptsize{$\pm$0.1} & 90.6\scriptsize{$\pm$0.2} & 92.0\scriptsize{$\pm$0.3} & 92.4\scriptsize{$\pm$0.2} & 95.8\scriptsize{$\pm$0.2} \\
    \bottomrule
  \end{tabular}
\end{table*}

\begin{table*}
  \centering
  \caption{Quantitative results of the \textbf{2-shot} setting on the \textbf{MVTec AD}~\cite{mvtecad} dataset. We report the mean and standard deviation over 5 random seeds for each measurement.}
  \label{tab:2_shot_mvtec}
  \tablestyle{6pt}{1.05}
  \setlength\tabcolsep{1.0pt}
  \begin{tabular}{@{}l p{1.8cm}<{\centering} p{1.8cm}<{\centering} p{1.8cm}<{\centering} p{1.8cm}<{\centering} p{1.8cm}<{\centering} p{1.8cm}<{\centering} p{1.8cm}<{\centering}@{}}
    \toprule
    Object & AUROC-segm & F1-max-segm & AP-segm & PRO-segm & AUROC-cls & F1-max-cls & AP-cls\\
    \midrule
    carpet & 98.7\scriptsize{$\pm$0.0} & 71.7\scriptsize{$\pm$0.1} & 71.7\scriptsize{$\pm$0.1} & 96.7\scriptsize{$\pm$0.1} & 99.9\scriptsize{$\pm$0.0} & 99.0\scriptsize{$\pm$0.2} & 100.0\scriptsize{$\pm$0.0} \\
    bottle & 96.9\scriptsize{$\pm$0.1} & 73.3\scriptsize{$\pm$0.7} & 75.6\scriptsize{$\pm$0.6} & 93.6\scriptsize{$\pm$0.2} & 94.0\scriptsize{$\pm$0.4} & 93.2\scriptsize{$\pm$0.4} & 98.2\scriptsize{$\pm$0.1} \\
    hazelnut & 97.7\scriptsize{$\pm$0.1} & 60.6\scriptsize{$\pm$0.3} & 60.8\scriptsize{$\pm$0.7} & 92.5\scriptsize{$\pm$0.2} & 98.7\scriptsize{$\pm$0.1} & 97.1\scriptsize{$\pm$0.4} & 99.4\scriptsize{$\pm$0.1} \\
    leather & 99.5\scriptsize{$\pm$0.0} & 51.5\scriptsize{$\pm$0.2} & 56.0\scriptsize{$\pm$0.1} & 99.0\scriptsize{$\pm$0.1} & 100.0\scriptsize{$\pm$0.0} & 99.8\scriptsize{$\pm$0.2} & 100.0\scriptsize{$\pm$0.0} \\
    cable & 91.2\scriptsize{$\pm$0.8} & 38.9\scriptsize{$\pm$0.9} & 33.1\scriptsize{$\pm$0.7} & 82.9\scriptsize{$\pm$0.8} & 74.9\scriptsize{$\pm$1.0} & 78.2\scriptsize{$\pm$0.9} & 85.1\scriptsize{$\pm$0.8} \\
    capsule & 97.1\scriptsize{$\pm$0.3} & 39.2\scriptsize{$\pm$4.9} & 38.2\scriptsize{$\pm$4.5} & 96.0\scriptsize{$\pm$0.9} & 92.1\scriptsize{$\pm$2.1} & 92.9\scriptsize{$\pm$0.5} & 98.4\scriptsize{$\pm$0.5} \\
    grid & 96.9\scriptsize{$\pm$0.2} & 48.7\scriptsize{$\pm$0.6} & 42.2\scriptsize{$\pm$0.4} & 90.9\scriptsize{$\pm$0.8} & 99.0\scriptsize{$\pm$0.1} & 97.7\scriptsize{$\pm$0.4} & 99.6\scriptsize{$\pm$0.0} \\
    pill & 95.0\scriptsize{$\pm$0.3} & 55.2\scriptsize{$\pm$0.8} & 47.0\scriptsize{$\pm$0.9} & 96.7\scriptsize{$\pm$0.2} & 84.1\scriptsize{$\pm$0.2} & 92.4\scriptsize{$\pm$0.2} & 96.9\scriptsize{$\pm$0.1} \\
    transistor & 82.5\scriptsize{$\pm$0.5} & 38.5\scriptsize{$\pm$1.5} & 27.9\scriptsize{$\pm$1.0} & 66.5\scriptsize{$\pm$0.6} & 83.8\scriptsize{$\pm$0.7} & 74.8\scriptsize{$\pm$1.8} & 77.5\scriptsize{$\pm$1.7} \\
    metal\_nut & 92.1\scriptsize{$\pm$0.5} & 62.5\scriptsize{$\pm$2.3} & 60.7\scriptsize{$\pm$2.0} & 90.5\scriptsize{$\pm$0.4} & 90.5\scriptsize{$\pm$0.3} & 91.9\scriptsize{$\pm$0.6} & 97.8\scriptsize{$\pm$0.0} \\
    screw & 98.3\scriptsize{$\pm$0.1} & 43.2\scriptsize{$\pm$0.4} & 36.2\scriptsize{$\pm$1.1} & 93.1\scriptsize{$\pm$0.6} & 82.0\scriptsize{$\pm$2.2} & 88.8\scriptsize{$\pm$0.3} & 92.4\scriptsize{$\pm$1.4} \\
    toothbrush & 98.5\scriptsize{$\pm$0.2} & 55.6\scriptsize{$\pm$0.7} & 55.9\scriptsize{$\pm$1.3} & 93.8\scriptsize{$\pm$0.5} & 94.8\scriptsize{$\pm$1.3} & 93.3\scriptsize{$\pm$0.5} & 98.0\scriptsize{$\pm$0.5} \\
    zipper & 96.4\scriptsize{$\pm$0.1} & 59.2\scriptsize{$\pm$0.7} & 57.2\scriptsize{$\pm$0.6} & 88.7\scriptsize{$\pm$0.5} & 94.7\scriptsize{$\pm$0.5} & 94.7\scriptsize{$\pm$0.5} & 98.4\scriptsize{$\pm$0.2} \\
    tile & 95.8\scriptsize{$\pm$0.1} & 72.3\scriptsize{$\pm$0.2} & 69.9\scriptsize{$\pm$0.3} & 92.3\scriptsize{$\pm$0.3} & 99.0\scriptsize{$\pm$0.1} & 97.7\scriptsize{$\pm$0.0} & 99.6\scriptsize{$\pm$0.0} \\
    wood & 96.1\scriptsize{$\pm$0.1} & 67.7\scriptsize{$\pm$0.3} & 68.8\scriptsize{$\pm$0.3} & 96.0\scriptsize{$\pm$0.1} & 98.7\scriptsize{$\pm$0.3} & 97.5\scriptsize{$\pm$0.0} & 99.6\scriptsize{$\pm$0.1} \\
    \midrule
    Mean & 95.5\scriptsize{$\pm$0.0} & 55.9\scriptsize{$\pm$0.5} & 53.4\scriptsize{$\pm$0.4} & 91.3\scriptsize{$\pm$0.1} & 92.4\scriptsize{$\pm$0.3} & 92.6\scriptsize{$\pm$0.1} & 96.0\scriptsize{$\pm$0.2} \\
    \bottomrule
  \end{tabular}
\end{table*}

\begin{table*}
  \centering
  \caption{Quantitative results of the \textbf{4-shot} setting on the \textbf{MVTec AD}~\cite{mvtecad} dataset. We report the mean and standard deviation over 5 random seeds for each measurement.}
  \label{tab:4_shot_mvtec}
  \tablestyle{6pt}{1.05}
  \setlength\tabcolsep{1.0pt}
  \begin{tabular}{@{}l p{1.8cm}<{\centering} p{1.8cm}<{\centering} p{1.8cm}<{\centering} p{1.8cm}<{\centering} p{1.8cm}<{\centering} p{1.8cm}<{\centering} p{1.8cm}<{\centering}@{}}
    \toprule
    Object & AUROC-segm & F1-max-segm & AP-segm & PRO-segm & AUROC-cls & F1-max-cls & AP-cls\\
    \midrule
    carpet & 98.7\scriptsize{$\pm$0.0} & 71.8\scriptsize{$\pm$0.1} & 71.8\scriptsize{$\pm$0.1} & 96.6\scriptsize{$\pm$0.2} & 99.9\scriptsize{$\pm$0.0} & 99.0\scriptsize{$\pm$0.2} & 100.0\scriptsize{$\pm$0.0} \\
    bottle & 97.2\scriptsize{$\pm$0.1} & 74.1\scriptsize{$\pm$0.6} & 76.5\scriptsize{$\pm$0.5} & 94.3\scriptsize{$\pm$0.3} & 94.2\scriptsize{$\pm$0.2} & 93.3\scriptsize{$\pm$0.4} & 98.3\scriptsize{$\pm$0.0} \\
    hazelnut & 97.7\scriptsize{$\pm$0.1} & 60.5\scriptsize{$\pm$0.4} & 60.7\scriptsize{$\pm$0.5} & 92.6\scriptsize{$\pm$0.2} & 98.8\scriptsize{$\pm$0.1} & 97.1\scriptsize{$\pm$0.4} & 99.4\scriptsize{$\pm$0.1} \\
    leather & 99.5\scriptsize{$\pm$0.0} & 51.5\scriptsize{$\pm$0.1} & 55.9\scriptsize{$\pm$0.1} & 98.9\scriptsize{$\pm$0.1} & 100.0\scriptsize{$\pm$0.0} & 99.9\scriptsize{$\pm$0.2} & 100.0\scriptsize{$\pm$0.0} \\
    cable & 91.8\scriptsize{$\pm$0.7} & 41.2\scriptsize{$\pm$0.7} & 34.5\scriptsize{$\pm$0.5} & 84.0\scriptsize{$\pm$0.8} & 76.7\scriptsize{$\pm$0.8} & 79.0\scriptsize{$\pm$0.5} & 86.5\scriptsize{$\pm$0.5} \\
    capsule & 97.5\scriptsize{$\pm$0.2} & 42.2\scriptsize{$\pm$0.6} & 41.0\scriptsize{$\pm$0.5} & 96.6\scriptsize{$\pm$0.3} & 93.5\scriptsize{$\pm$0.8} & 93.3\scriptsize{$\pm$0.6} & 98.7\scriptsize{$\pm$0.2} \\
    grid & 97.6\scriptsize{$\pm$0.3} & 49.0\scriptsize{$\pm$0.9} & 43.4\scriptsize{$\pm$1.0} & 92.0\scriptsize{$\pm$0.8} & 99.2\scriptsize{$\pm$0.2} & 98.1\scriptsize{$\pm$0.3} & 99.7\scriptsize{$\pm$0.1} \\
    pill & 95.5\scriptsize{$\pm$0.2} & 56.3\scriptsize{$\pm$0.2} & 48.2\scriptsize{$\pm$0.5} & 96.7\scriptsize{$\pm$0.1} & 84.1\scriptsize{$\pm$0.9} & 92.3\scriptsize{$\pm$0.1} & 96.9\scriptsize{$\pm$0.2} \\
    transistor & 83.7\scriptsize{$\pm$1.0} & 40.0\scriptsize{$\pm$2.2} & 29.2\scriptsize{$\pm$1.8} & 67.7\scriptsize{$\pm$1.4} & 84.1\scriptsize{$\pm$1.0} & 74.4\scriptsize{$\pm$0.5} & 78.1\scriptsize{$\pm$1.5} \\
    metal\_nut & 93.1\scriptsize{$\pm$0.7} & 66.7\scriptsize{$\pm$3.0} & 64.2\scriptsize{$\pm$2.4} & 91.7\scriptsize{$\pm$0.7} & 91.0\scriptsize{$\pm$0.4} & 92.2\scriptsize{$\pm$0.6} & 97.9\scriptsize{$\pm$0.1} \\
    screw & 98.5\scriptsize{$\pm$0.2} & 43.7\scriptsize{$\pm$1.2} & 37.7\scriptsize{$\pm$2.2} & 93.7\scriptsize{$\pm$0.3} & 83.7\scriptsize{$\pm$2.3} & 89.5\scriptsize{$\pm$0.7} & 93.2\scriptsize{$\pm$1.7} \\
    toothbrush & 98.8\scriptsize{$\pm$0.1} & 55.7\scriptsize{$\pm$0.8} & 56.8\scriptsize{$\pm$1.0} & 94.8\scriptsize{$\pm$0.9} & 93.2\scriptsize{$\pm$0.7} & 92.6\scriptsize{$\pm$0.6} & 97.3\scriptsize{$\pm$0.3} \\
    zipper & 96.6\scriptsize{$\pm$0.1} & 60.2\scriptsize{$\pm$1.0} & 58.2\scriptsize{$\pm$0.9} & 89.2\scriptsize{$\pm$0.5} & 95.4\scriptsize{$\pm$0.4} & 95.7\scriptsize{$\pm$0.8} & 98.6\scriptsize{$\pm$0.1} \\
    tile & 96.0\scriptsize{$\pm$0.1} & 72.5\scriptsize{$\pm$0.2} & 70.2\scriptsize{$\pm$0.2} & 92.4\scriptsize{$\pm$0.2} & 99.1\scriptsize{$\pm$0.0} & 97.8\scriptsize{$\pm$0.2} & 99.6\scriptsize{$\pm$0.0} \\
    wood & 96.2\scriptsize{$\pm$0.0} & 67.7\scriptsize{$\pm$0.2} & 69.0\scriptsize{$\pm$0.2} & 96.1\scriptsize{$\pm$0.2} & 98.7\scriptsize{$\pm$0.2} & 97.5\scriptsize{$\pm$0.0} & 99.6\scriptsize{$\pm$0.1} \\
    \midrule
    Mean & 95.9\scriptsize{$\pm$0.0} & 56.9\scriptsize{$\pm$0.1} & 54.5\scriptsize{$\pm$0.2} & 91.8\scriptsize{$\pm$0.1} & 92.8\scriptsize{$\pm$0.2} & 92.8\scriptsize{$\pm$0.1} & 96.3\scriptsize{$\pm$0.1} \\
    \bottomrule
  \end{tabular}
\end{table*}

\begin{table*}
  \centering
  \caption{Quantitative results of the \textbf{zero-shot} setting on the \textbf{VisA}~\cite{visa} dataset.}
  \label{tab:zero_shot_visa}
  \tablestyle{6pt}{1.05}
  \setlength\tabcolsep{1.0pt}
  \begin{tabular}{@{}l p{1.1cm}<{\centering} p{1.1cm}<{\centering} p{1.1cm}<{\centering} p{1.1cm}<{\centering} p{1.1cm}<{\centering} p{1.1cm}<{\centering} p{1.1cm}<{\centering}@{}}
    \toprule
    Object & AUROC-segm & F1-max-segm & AP-segm & PRO-segm & AUROC-cls & F1-max-cls & AP-cls\\
    \midrule
    candle&97.8&39.4&29.9&92.5&83.8&77.8&86.9\\
    capsules&97.5&48.5&40.0&86.7&61.2&77.6&74.3\\
    cashew&86.0&22.9&15.1&91.7&87.3&84.8&94.1\\
    chewinggum&99.5&78.5&83.6&87.3&96.4&93.7&98.4\\
    fryum&92.0&29.7&22.1&89.7&94.3&91.7&97.2\\
    macaroni1&98.8&35.5&24.8&93.2&71.6&71.1&70.9\\
    macaroni2&97.8&13.7&6.8&82.3&64.6&69.1&63.2\\
    pcb1&92.7&12.5&8.4&87.5&53.4&66.9&57.2\\
    pcb2&89.7&23.4&15.4&75.6&71.8&70.1&73.8\\
    pcb3&88.4&21.7&14.1&77.8&66.8&66.7&70.7\\
    pcb4&94.6&31.3&24.9&86.8&95.0&87.3&95.1\\
    pipe\_fryum&96.0&30.4&23.6&90.9&89.9&87.7&94.8\\
    \midrule
    Mean&94.2&32.3&25.7&86.8&78.0&78.7&81.4\\
    \bottomrule
  \end{tabular}
\end{table*}

\begin{table*}
  \centering
  \caption{Quantitative results of the \textbf{1-shot} setting on the \textbf{VisA}~\cite{visa} dataset. We report the mean and standard deviation over 5 random seeds for each measurement.}
  \label{tab:1_shot_visa}
  \tablestyle{6pt}{1.05}
  \setlength\tabcolsep{1.0pt}
  \begin{tabular}{@{}l p{1.8cm}<{\centering} p{1.8cm}<{\centering} p{1.8cm}<{\centering} p{1.8cm}<{\centering} p{1.8cm}<{\centering} p{1.8cm}<{\centering} p{1.8cm}<{\centering}@{}}
    \toprule
    Object & AUROC-segm & F1-max-segm & AP-segm & PRO-segm & AUROC-cls & F1-max-cls & AP-cls\\
    \midrule
    candle & 98.7\scriptsize{$\pm$0.1} & 42.4\scriptsize{$\pm$0.3} & 29.7\scriptsize{$\pm$1.4} & 96.4\scriptsize{$\pm$0.2} & 90.9\scriptsize{$\pm$0.5} & 84.3\scriptsize{$\pm$0.9} & 91.6\scriptsize{$\pm$0.7} \\
    capsules & 98.0\scriptsize{$\pm$0.0} & 51.8\scriptsize{$\pm$0.2} & 45.3\scriptsize{$\pm$0.4} & 88.4\scriptsize{$\pm$0.1} & 92.7\scriptsize{$\pm$1.0} & 90.2\scriptsize{$\pm$1.0} & 95.8\scriptsize{$\pm$0.7} \\
    cashew & 90.8\scriptsize{$\pm$0.1} & 30.9\scriptsize{$\pm$0.5} & 23.1\scriptsize{$\pm$0.5} & 94.2\scriptsize{$\pm$0.2} & 93.4\scriptsize{$\pm$1.0} & 90.7\scriptsize{$\pm$0.9} & 97.1\scriptsize{$\pm$0.5} \\
    chewinggum & 99.7\scriptsize{$\pm$0.0} & 78.8\scriptsize{$\pm$0.3} & 82.2\scriptsize{$\pm$0.5} & 92.1\scriptsize{$\pm$0.2} & 97.3\scriptsize{$\pm$0.1} & 97.1\scriptsize{$\pm$0.4} & 99.0\scriptsize{$\pm$0.0} \\
    fryum & 93.6\scriptsize{$\pm$0.1} & 33.5\scriptsize{$\pm$0.2} & 25.9\scriptsize{$\pm$0.2} & 91.5\scriptsize{$\pm$0.3} & 93.4\scriptsize{$\pm$0.7} & 91.9\scriptsize{$\pm$0.8} & 97.3\scriptsize{$\pm$0.3} \\
    macaroni1 & 99.3\scriptsize{$\pm$0.0} & 37.2\scriptsize{$\pm$0.2} & 26.5\scriptsize{$\pm$0.5} & 95.3\scriptsize{$\pm$0.1} & 89.7\scriptsize{$\pm$0.6} & 82.5\scriptsize{$\pm$0.8} & 92.0\scriptsize{$\pm$0.4} \\
    macaroni2 & 98.4\scriptsize{$\pm$0.1} & 22.5\scriptsize{$\pm$1.9} & 11.5\scriptsize{$\pm$1.2} & 85.6\scriptsize{$\pm$0.7} & 79.2\scriptsize{$\pm$3.8} & 72.9\scriptsize{$\pm$2.2} & 83.4\scriptsize{$\pm$3.2} \\
    pcb1 & 95.3\scriptsize{$\pm$0.1} & 20.6\scriptsize{$\pm$1.6} & 15.8\scriptsize{$\pm$0.8} & 90.4\scriptsize{$\pm$0.6} & 89.0\scriptsize{$\pm$7.3} & 81.6\scriptsize{$\pm$6.0} & 90.2\scriptsize{$\pm$5.5} \\
    pcb2 & 92.5\scriptsize{$\pm$0.1} & 31.5\scriptsize{$\pm$0.9} & 20.5\scriptsize{$\pm$1.4} & 79.2\scriptsize{$\pm$0.3} & 85.4\scriptsize{$\pm$6.5} & 79.8\scriptsize{$\pm$3.5} & 87.1\scriptsize{$\pm$7.8} \\
    pcb3 & 93.2\scriptsize{$\pm$0.0} & 38.5\scriptsize{$\pm$1.1} & 28.1\scriptsize{$\pm$1.7} & 81.3\scriptsize{$\pm$0.2} & 87.8\scriptsize{$\pm$1.2} & 82.0\scriptsize{$\pm$1.3} & 89.4\scriptsize{$\pm$1.2} \\
    pcb4 & 95.8\scriptsize{$\pm$0.1} & 37.6\scriptsize{$\pm$0.9} & 32.6\scriptsize{$\pm$1.3} & 89.3\scriptsize{$\pm$0.3} & 97.5\scriptsize{$\pm$0.3} & 93.7\scriptsize{$\pm$0.7} & 97.0\scriptsize{$\pm$0.5} \\
    pipe\_fryum & 97.3\scriptsize{$\pm$0.0} & 36.6\scriptsize{$\pm$0.3} & 29.6\scriptsize{$\pm$0.5} & 96.1\scriptsize{$\pm$0.1} & 98.3\scriptsize{$\pm$0.4} & 96.3\scriptsize{$\pm$0.6} & 99.2\scriptsize{$\pm$0.2} \\
    \midrule
    Mean & 96.0\scriptsize{$\pm$0.0} & 38.5\scriptsize{$\pm$0.3} & 30.9\scriptsize{$\pm$0.3} & 90.0\scriptsize{$\pm$0.1} & 91.2\scriptsize{$\pm$0.8} & 86.9\scriptsize{$\pm$0.6} & 93.3\scriptsize{$\pm$0.8} \\
    \bottomrule
  \end{tabular}
\end{table*}

\begin{table*}
  \centering
  \caption{Quantitative results of the \textbf{2-shot} setting on the \textbf{VisA}~\cite{visa} dataset. We report the mean and standard deviation over 5 random seeds for each measurement.}
  \label{tab:2_shot_visa}
  \tablestyle{6pt}{1.05}
  \setlength\tabcolsep{1.0pt}
  \begin{tabular}{@{}l p{1.8cm}<{\centering} p{1.8cm}<{\centering} p{1.8cm}<{\centering} p{1.8cm}<{\centering} p{1.8cm}<{\centering} p{1.8cm}<{\centering} p{1.8cm}<{\centering}@{}}
    \toprule
    Object & AUROC-segm & F1-max-segm & AP-segm & PRO-segm & AUROC-cls & F1-max-cls & AP-cls\\
    \midrule
    candle & 98.6\scriptsize{$\pm$0.1} & 42.4\scriptsize{$\pm$0.3} & 29.5\scriptsize{$\pm$1.6} & 96.5\scriptsize{$\pm$0.2} & 91.3\scriptsize{$\pm$0.5} & 84.3\scriptsize{$\pm$0.4} & 91.9\scriptsize{$\pm$0.6} \\
    capsules & 98.0\scriptsize{$\pm$0.0} & 52.2\scriptsize{$\pm$0.2} & 45.7\scriptsize{$\pm$0.3} & 89.1\scriptsize{$\pm$0.3} & 93.4\scriptsize{$\pm$0.3} & 90.4\scriptsize{$\pm$0.7} & 96.3\scriptsize{$\pm$0.2} \\
    cashew & 90.8\scriptsize{$\pm$0.1} & 31.4\scriptsize{$\pm$0.4} & 23.4\scriptsize{$\pm$0.5} & 94.1\scriptsize{$\pm$0.2} & 93.4\scriptsize{$\pm$0.6} & 91.6\scriptsize{$\pm$0.5} & 97.2\scriptsize{$\pm$0.3} \\
    chewinggum & 99.7\scriptsize{$\pm$0.0} & 78.8\scriptsize{$\pm$0.3} & 82.4\scriptsize{$\pm$0.3} & 92.1\scriptsize{$\pm$0.3} & 97.1\scriptsize{$\pm$0.1} & 96.9\scriptsize{$\pm$0.4} & 98.9\scriptsize{$\pm$0.0} \\
    fryum & 93.6\scriptsize{$\pm$0.0} & 34.0\scriptsize{$\pm$0.1} & 26.3\scriptsize{$\pm$0.1} & 91.6\scriptsize{$\pm$0.1} & 93.1\scriptsize{$\pm$0.4} & 91.5\scriptsize{$\pm$0.7} & 97.2\scriptsize{$\pm$0.1} \\
    macaroni1 & 99.3\scriptsize{$\pm$0.0} & 37.0\scriptsize{$\pm$0.4} & 26.3\scriptsize{$\pm$0.7} & 95.2\scriptsize{$\pm$0.1} & 90.0\scriptsize{$\pm$0.4} & 82.5\scriptsize{$\pm$0.4} & 92.1\scriptsize{$\pm$0.2} \\
    macaroni2 & 98.4\scriptsize{$\pm$0.0} & 24.2\scriptsize{$\pm$0.8} & 12.3\scriptsize{$\pm$0.6} & 85.4\scriptsize{$\pm$0.2} & 81.0\scriptsize{$\pm$1.0} & 74.1\scriptsize{$\pm$1.1} & 85.1\scriptsize{$\pm$0.9} \\
    pcb1 & 95.8\scriptsize{$\pm$0.2} & 24.2\scriptsize{$\pm$3.2} & 18.9\scriptsize{$\pm$2.4} & 90.6\scriptsize{$\pm$0.2} & 90.9\scriptsize{$\pm$1.3} & 83.8\scriptsize{$\pm$0.8} & 91.6\scriptsize{$\pm$0.9} \\
    pcb2 & 92.8\scriptsize{$\pm$0.1} & 32.6\scriptsize{$\pm$0.4} & 21.6\scriptsize{$\pm$0.3} & 79.7\scriptsize{$\pm$0.3} & 89.2\scriptsize{$\pm$1.5} & 83.3\scriptsize{$\pm$1.7} & 91.3\scriptsize{$\pm$1.2} \\
    pcb3 & 93.5\scriptsize{$\pm$0.0} & 40.5\scriptsize{$\pm$0.8} & 30.6\scriptsize{$\pm$1.0} & 81.5\scriptsize{$\pm$0.5} & 90.9\scriptsize{$\pm$0.8} & 84.5\scriptsize{$\pm$1.2} & 92.2\scriptsize{$\pm$0.7} \\
    pcb4 & 95.9\scriptsize{$\pm$0.1} & 38.3\scriptsize{$\pm$0.5} & 33.0\scriptsize{$\pm$0.8} & 89.5\scriptsize{$\pm$0.3} & 97.9\scriptsize{$\pm$0.5} & 93.1\scriptsize{$\pm$1.1} & 97.8\scriptsize{$\pm$0.6} \\
    pipe\_fryum & 97.3\scriptsize{$\pm$0.0} & 36.8\scriptsize{$\pm$0.2} & 29.8\scriptsize{$\pm$0.3} & 96.1\scriptsize{$\pm$0.3} & 98.4\scriptsize{$\pm$0.2} & 96.2\scriptsize{$\pm$0.6} & 99.3\scriptsize{$\pm$0.1} \\
    \midrule
    Mean & 96.2\scriptsize{$\pm$0.0} & 39.3\scriptsize{$\pm$0.2} & 31.6\scriptsize{$\pm$0.3} & 90.1\scriptsize{$\pm$0.1} & 92.2\scriptsize{$\pm$0.3} & 87.7\scriptsize{$\pm$0.3} & 94.2\scriptsize{$\pm$0.3} \\
    \bottomrule
  \end{tabular}
\end{table*}

\begin{table*}
  \centering
  \caption{Quantitative results of the \textbf{4-shot} setting on the \textbf{VisA}~\cite{visa} dataset. We report the mean and standard deviation over 5 random seeds for each measurement.}
  \label{tab:4_shot_visa}
  \tablestyle{6pt}{1.05}
  \setlength\tabcolsep{1.0pt}
  \begin{tabular}{@{}l p{1.8cm}<{\centering} p{1.8cm}<{\centering} p{1.8cm}<{\centering} p{1.8cm}<{\centering} p{1.8cm}<{\centering} p{1.8cm}<{\centering} p{1.8cm}<{\centering}@{}}
    \toprule
    Object & AUROC-segm & F1-max-segm & AP-segm & PRO-segm & AUROC-cls & F1-max-cls & AP-cls\\
    \midrule
    candle & 98.7\scriptsize{$\pm$0.0} & 42.4\scriptsize{$\pm$0.2} & 29.6\scriptsize{$\pm$1.0} & 96.3\scriptsize{$\pm$0.1} & 91.4\scriptsize{$\pm$0.5} & 85.1\scriptsize{$\pm$0.5} & 92.0\scriptsize{$\pm$0.6} \\
    capsules & 98.1\scriptsize{$\pm$0.0} & 52.4\scriptsize{$\pm$0.3} & 46.0\scriptsize{$\pm$0.5} & 89.0\scriptsize{$\pm$0.4} & 93.7\scriptsize{$\pm$0.1} & 90.6\scriptsize{$\pm$0.4} & 96.5\scriptsize{$\pm$0.1} \\
    cashew & 90.8\scriptsize{$\pm$0.1} & 31.6\scriptsize{$\pm$0.3} & 23.7\scriptsize{$\pm$0.3} & 94.1\scriptsize{$\pm$0.2} & 94.3\scriptsize{$\pm$0.3} & 92.2\scriptsize{$\pm$0.5} & 97.6\scriptsize{$\pm$0.2} \\
    chewinggum & 99.7\scriptsize{$\pm$0.0} & 78.8\scriptsize{$\pm$0.3} & 82.0\scriptsize{$\pm$0.2} & 92.2\scriptsize{$\pm$0.3} & 97.2\scriptsize{$\pm$0.2} & 97.2\scriptsize{$\pm$0.2} & 99.0\scriptsize{$\pm$0.0} \\
    fryum & 93.7\scriptsize{$\pm$0.0} & 34.2\scriptsize{$\pm$0.0} & 26.4\scriptsize{$\pm$0.0} & 91.5\scriptsize{$\pm$0.1} & 93.5\scriptsize{$\pm$0.5} & 92.3\scriptsize{$\pm$1.2} & 97.4\scriptsize{$\pm$0.2} \\
    macaroni1 & 99.3\scriptsize{$\pm$0.0} & 37.3\scriptsize{$\pm$0.4} & 27.2\scriptsize{$\pm$0.5} & 95.2\scriptsize{$\pm$0.1} & 90.1\scriptsize{$\pm$0.4} & 82.7\scriptsize{$\pm$0.7} & 92.4\scriptsize{$\pm$0.3} \\
    macaroni2 & 98.4\scriptsize{$\pm$0.0} & 24.6\scriptsize{$\pm$0.4} & 12.5\scriptsize{$\pm$0.4} & 85.8\scriptsize{$\pm$0.2} & 82.5\scriptsize{$\pm$0.5} & 76.2\scriptsize{$\pm$0.8} & 86.3\scriptsize{$\pm$0.3} \\
    pcb1 & 96.0\scriptsize{$\pm$0.0} & 27.7\scriptsize{$\pm$1.1} & 21.7\scriptsize{$\pm$0.7} & 90.6\scriptsize{$\pm$0.1} & 91.2\scriptsize{$\pm$1.5} & 85.2\scriptsize{$\pm$1.8} & 91.8\scriptsize{$\pm$1.0} \\
    pcb2 & 93.0\scriptsize{$\pm$0.1} & 32.4\scriptsize{$\pm$0.6} & 21.8\scriptsize{$\pm$0.4} & 80.2\scriptsize{$\pm$0.3} & 89.2\scriptsize{$\pm$0.7} & 82.6\scriptsize{$\pm$1.7} & 91.3\scriptsize{$\pm$0.7} \\
    pcb3 & 93.7\scriptsize{$\pm$0.1} & 41.9\scriptsize{$\pm$0.4} & 31.4\scriptsize{$\pm$0.6} & 81.5\scriptsize{$\pm$0.3} & 91.6\scriptsize{$\pm$0.4} & 85.7\scriptsize{$\pm$0.7} & 92.7\scriptsize{$\pm$0.3} \\
    pcb4 & 96.0\scriptsize{$\pm$0.0} & 39.2\scriptsize{$\pm$0.3} & 34.1\scriptsize{$\pm$0.3} & 89.7\scriptsize{$\pm$0.1} & 98.2\scriptsize{$\pm$0.3} & 94.1\scriptsize{$\pm$0.3} & 98.0\scriptsize{$\pm$0.3} \\
    pipe\_fryum & 97.4\scriptsize{$\pm$0.0} & 36.8\scriptsize{$\pm$0.3} & 30.0\scriptsize{$\pm$0.3} & 95.9\scriptsize{$\pm$0.2} & 98.5\scriptsize{$\pm$0.4} & 96.8\scriptsize{$\pm$0.7} & 99.4\scriptsize{$\pm$0.2} \\
    \midrule
    Mean & 96.2\scriptsize{$\pm$0.0} & 40.0\scriptsize{$\pm$0.1} & 32.2\scriptsize{$\pm$0.1} & 90.2\scriptsize{$\pm$0.1} & 92.6\scriptsize{$\pm$0.4} & 88.4\scriptsize{$\pm$0.5} & 94.5\scriptsize{$\pm$0.3} \\
    \bottomrule
  \end{tabular}
\end{table*}

{\small
\bibliographystyle{ieee_fullname}
\bibliography{arxiv}
}

\end{document}